 \documentclass[pmlr,twocolumn,10pt]{jmlr} 

\usepackage{booktabs}
\usepackage{siunitx}
\usepackage{multirow}

\usepackage[switch]{lineno}

\theorembodyfont{\upshape}
\theoremheaderfont{\scshape}
\theorempostheader{:}
\theoremsep{\newline}

\jmlrvolume{LEAVE UNSET}
\jmlryear{2024}
\jmlrsubmitted{LEAVE UNSET}
\jmlrpublished{LEAVE UNSET}
\jmlrworkshop{Machine Learning for Health (ML4H) 2024} 

\title[Thermography Feasibility for Pressure Injury Detection in Dark Skin Patients]{Is thermography a viable solution for detecting pressure injuries in dark skin patients?}

\author{%
 \Name{Miriam Asare-Baiden} \Email{masareb@emory.edu}\\
 \Name{Kathleen Jordan} \Email{k.a.jordan@emory.edu}\\
 \Name{Andrew Chung} \Email{chun46@emory.edu}\\
\Name{Sharon {Eve Sonenblum}} \Email{sharoneve@emory.edu}\\
\Name{Joyce {C. Ho}} \Email{joyce.c.ho@emory.edu}\\
\addr Emory University, USA
}

\begin{document}

\maketitle

\begin{abstract}
Pressure injury (PI) detection is challenging, especially in dark skin tones, due to the unreliability of visual inspection. Thermography has been suggested as a viable alternative as temperature differences in the skin can indicate impending tissue damage. Although deep learning models have demonstrated considerable promise toward reliably detecting PI, the existing work fails to evaluate the performance on darker skin tones and varying data collection protocols. In this paper, we introduce a new thermal and optical imaging dataset of 35 participants focused on darker skin tones where temperature differences are induced through cooling and cupping protocols. We vary the image collection process to include different cameras, lighting, patient pose, and camera distance. We compare the performance of a small convolutional neural network (CNN) trained on either the thermal or the optical images on all skin tones.
Our preliminary results suggest that thermography-based CNN is robust to data collection protocols for all skin tones. 
\end{abstract}




\section{Introduction}
\label{sec:intro}
A pressure injury (PI) refers to ``localized damage to the skin and underlying soft tissue, usually occurring over a bony prominence or related to medical devices" \citep{epuap_npuap_pppia_2019}. 
It continues to be a major issue with at least 1 in 10 adults admitted to the hospital developing a PI \citep{li2020global}.
Furthermore, PI is linked to a decline in quality of life \citep{khor2014determinants}, increased mortality rates \citep{bergquist2013pressure}, extended hospital stays, and a higher likelihood of requiring institutional care after discharge. 
Identifying warning signs of PI, such as erythema \citep{shi2020nonblanchable,guihan2012assessing}, can lead to effective preventive actions and early interventions \citep{jiang2020skin,baron2022efficacy}. 

Contemporary techniques for assessing PI risk include using the Braden Scale \citep{bergstrom1987braden,borghardt2015evaluation} and visual inspection. Visual inspection includes assessing the skin for temperature variations, changes in tissue consistency, and the presence of tenderness.
Yet visual inspection for dark skin tone patients is especially challenging and has led to racial disparities in PI outcomes \citep{black2023current}.
Given that PI may result in temperature abnormalities from partial or complete capillary occlusion at the site, thermography (i.e., thermal imaging) has been explored as an alternative technological approach for PI detection  \citep{baron2023accuracy}.
Several existing works have suggested deep learning models such as convolutional neural networks (CNNs) can achieve high predictive performance for detecting PI \citep{wang2021infrared,pandey2022deep,fergus2023pressure}.
However, these studies suffer from 3 major limitations.
(1) Limited work has been done to validate the use of thermal imaging across different skin tones \citep{aloweni2019evaluation,sprigle2003analysis}. 
There still remains an open question of how much melanin impacts the effectiveness of thermography by absorbing more infrared radiation.
For instance, infrared thermometers under-reported fevers in African Americans admitted to the hospital \citep{bhavani2022racial}. However, thermography sensitivity to melanin has remained understudied in the context of erythema and PI detection.
(2) There has been no work to compare the effectiveness of thermal imaging and optical imaging.
(3) Existing work trained on images acquired under strict data collection protocols where lighting, camera distance, and camera angle have been fixed. However, this is not practical in real-world practice.

\begin{figure*}[htbp]
\floatconts
{fig:image_example}
{\vspace{-18pt}\caption{Example of control images from our dataset with the 4 different skin tones.}}
{%
\subfigure[Optical Images]{%
\label{fig:skin-tones-optical}
\includegraphics[width=0.97\textwidth]{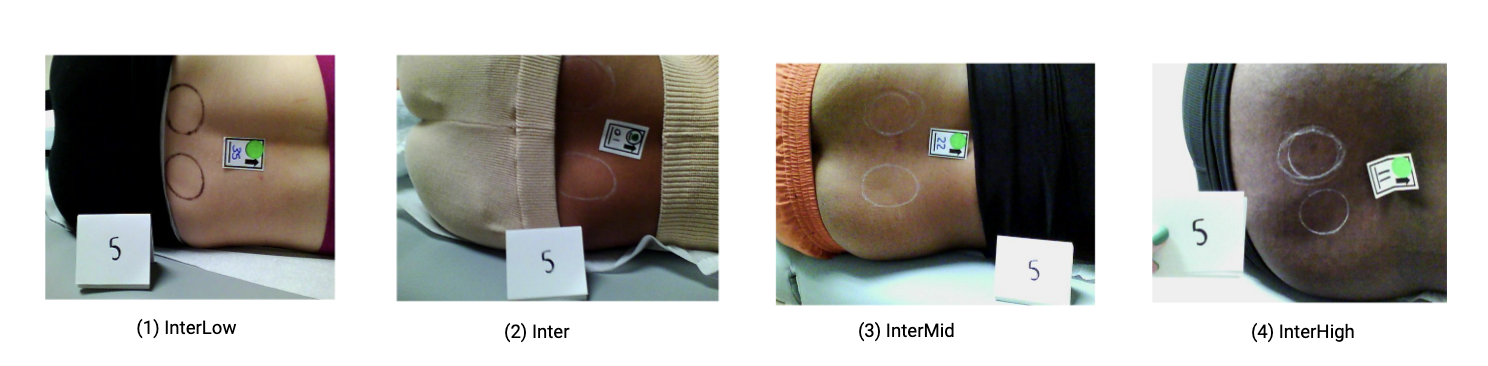}
}\qquad 
\subfigure[Thermal Images]{%
\label{fig:pic2}
\includegraphics[width=1\textwidth]{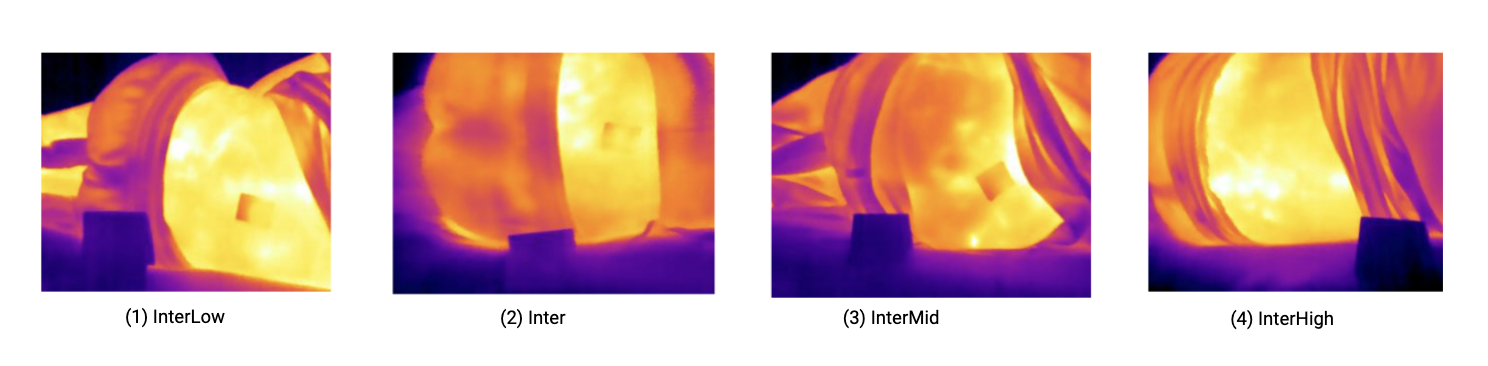}
}
}
\end{figure*}

The objective of our study is to address existing limitations by investigating the feasibility of thermography for detecting temperature differences in individuals with dark skin tones across various patient positions, lighting conditions, camera distance, and camera resolution.
We conducted an IRB-approved study to collect optical and thermal images from 35 healthy adults (predominantly darker skin tones) in a controlled clinical simulation environment and induced temperature differences on the lower back.
The controlled setup allows us to isolate temperature changes and assess whether thermography can reliably detect such changes, thus serving as a proxy for early PI detection.
We benchmark a convolutional neural network (CNN) model using our dataset and evaluate the efficacy of detecting these temperature changes across diverse dark skin tones. 
The key contributions of our research are as follows:
(1) Compilation of a diverse dark skin tone dataset comprising optical and thermal images from 35 patients;
(2) Comparison of a pre-trained CNN model's capability to detect temperature changes using either the optical or thermal images; and
(3) Analysis of how the imaging protocol can impact the model performance in a controlled clinical simulation environment.
Our findings are crucial to identifying optimal data collection approaches for real-world clinical settings.
\section{Method}

\subsection{Data Collection}
\label{sec:data_collection}
Our study collected optical and thermal images from 35 healthy adults in a controlled simulation environment. The study procedure was reviewed and approved by the Emory University Institutional Review Board, eRIB number 00005999. 30 participants had darker skin tones (Monk Skin Tone Scale level 6 or greater) and 5 had lighter skin tones (Monk Skin Tone Scale level 5 or lower). Using a digital colorimeter, we categorized the skin tone of subjects using Eumelanin Human Skin Colour Scale categories introduced by \citet{dadzie2022eumelanin} and revised by \citet{sonenblum2023using}. We had participants in the Eumelanin Intermediate Low (InterLow), Eumelanin Intermediate (Inter), Eumelanin Intermediate Mid (InterMid), and Eumelanin Intermediate High (InterHigh) categories as shown in Figure \ref{fig:image_example}. The lower back of each participant was marked to conduct both a cooling and cupping protocol separately. The cooling protocol involved a cooled stone cylinder placed on the right lower back. The cupping protocol was performed using a suction device to induce erythema (i.e., abnormal redness of the skin).

For the cooling protocol, images were taken with 2 thermal imaging cameras, the FLIR E8-XT (320x240 resolution) and the FLIR Pro One (160x120 resolution). 
The imaging collection protocol was varied to capture (i) two lighting settings (ambient and ring light), (ii) two distances (35 and 50 cm), and (iii) three different postures (forward placement of top knee, knees stacked, backward placement of top knee).
For each image acquisition setting (lighting, distance, posture) and camera, we captured a control and cool image pair (i.e., one before the cooling protocol and one after applying the stone cylinder).
This yielded a total of 1680 images for the cooling dataset.

For the cupping protocol, we used only the FLIR E8-XT at a single acquisition setting (50 cm and knees stacked).
Although the erythema task offers valuable insight, as visual indicators of erythema are often missed in individuals with darker skin, environmental factors such as lighting changes or camera distance adjustments can affect erythema visibility. We also observed erythema faded quickly in some subjects.
Moreover, some individuals may not visibly develop erythema at all. These challenges led to the restriction of testing thermography's effectiveness across varying conditions to only the cooling protocol.
We collected a control image and 8 images within the first 7 minutes after the cupping device was removed (9 images per patient).
Thus, we captured 314 images in the erythema dataset.\footnote{There is one missing due to a laptop failure.} 




\subsection{CNN}
CNNs have achieved remarkable performance in various medical imaging applications including dermatological assessments \citep{esteva2017dermatologist}.
\citet{wang2021infrared} proposed a CNN approach for classifying infrared thermal images, using 246 images from 82 patients with stringent imaging protocols. 
\citet{pandey2022deep} used the MobileNetV2 Object Detection Model \citep{giron2020classification} to identify the PI location using thermal images from 10 subjects and 18 images using a tripod, specific distance, and dietary restrictions. 
Faster RCNN was also used to categorize the PI stage for 849 optical images \citep{fergus2023pressure}. 
Though these CNN models showed remarkable results, model generalizability may not be assured due to the small dataset sizes.
 

We used MobileNetV2 which typically consists of an initial fully convolutional layer with 32 filters, followed by 19 residual bottleneck layers, and concludes with a 1x1 convolution, global average pooling, and a final fully connected layer \citep{sandler2018mobilenetv2}. The model's efficiency and effectiveness make it a promising candidate for real-time PI detection, particularly in scenarios where computational resources may be limited, such as in portable devices used for bedside assessments \citep{howard2019searching}.

\subsection*{Task Classification}
We evaluate the efficacy of 3 image types for detecting temperature: an optical image, a black and white (B\&W) thermal image and thermal color image. Thermal color images are the original temperatures captured directly by the thermal camera whereas  the  B\&W thermal images are grayscale version (0-255) derived from the color image. 
We assess the 3 image types on 2 tasks, cooling and erythema.
Both of these are binary classification tasks.
For the cooling task, we trained the model to identify whether the image was cool (+) or control (-). 
Similarly, for the erythema task, we classified the images as either positive if the measured erythema index is 6 c.u. above baseline (+), as measured with the colorimeter,  or control (-).
We randomly split the images into 80\% training and 20\% testing and ensured the image splits were the same across the 3 image types for fair comparison.
We evaluated performance on the test set using F1 and area under the receiver operating characteristic curve (AUC).

\subsection*{CNN Hyperparameter Tuning}
All images are resized to 224x224. We use 5-fold cross-validation to fine-tune a pre-trained MobileNetV2 on ImageNet for each image type and task. We employed data augmentation to supplement our training set by incorporating either a horizontal flip or a 20-degree rotation. We fine-tuned the MobileNetV2 with a batch size of 32, 50 epochs, a fixed learning rate, the Adam optimizer, and early stopping criteria.
The model that achieved the highest validation accuracy over 5-fold cross-validation was selected and used to classify the images in the test set. Additional details can be found in Appendix \ref{apd:methods}.

\section*{Results}

\begin{table}[t]
\floatconts
  {tab:mobilenet_results}%
  {\caption{Comparing performance across 2 tasks and image modalities for MobileNetV2.}}%
{\vspace{-20pt} \footnotesize
  \begin{tabular}{l l l l l }
\toprule
\textbf{Task} & \textbf{Image} & F1 & AUC \\
\midrule
 & Optical & 0.711 & 0.818 \\
Cooling & Thermal (B\&W) & 1.000 & 1.000 \\
& Thermal (Color) & 1.000 & 1.000 \\
\midrule
& Optical  & 0.868 & 0.909 \\
Erythema  & Thermal (B\&W)  & 0.914 & 0.938 \\
& Thermal (Color) & 0.918 & 0.935 \\
\bottomrule
\end{tabular}}
\end{table}

\subsection*{Overall Predictive Performance}
Table \ref{tab:mobilenet_results} summarizes the performance of MobileNetV2 on the 3 image types for the 2 tasks.
For both tasks, thermal imaging (B\&W and color) outperforms their optical counterparts.
For the cooling task (an easier and more obvious temperature difference task), both thermal image types achieve perfect performance with an AUC and F1 of 1, as opposed to 0.818 and 0.711 respectively for optical images.
For the erythema task, the optical image achieved better performance than the cooling task with an F1 and AUC of 0.858 and 0.909, respectively. This may be because erythema is accompanied by more visual color change than temperature change. The thermal-based CNN model achieves better performance with F1 and AUC scores of at least 0.914 and 0.935 respectively.
These results suggest thermography is more reliable than optical imaging for detecting temperature changes in darker skin tones.
Additional results for both tasks are available in Appendix \ref{apd:results}.

\subsection*{Impact of Image Protocol}
\begin{figure}[t]
    \centering
\includegraphics[width=0.89\linewidth]{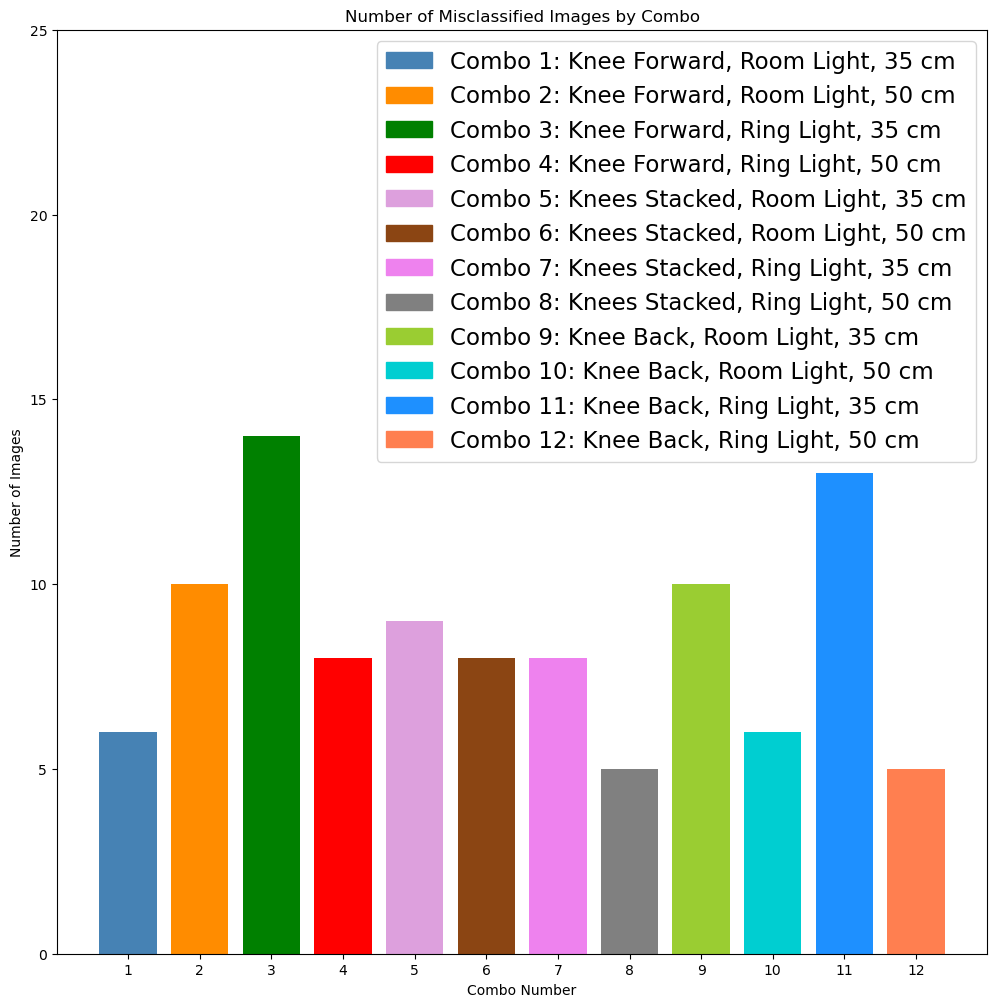}
\vspace{-12pt}
    \caption{Bar plot of misclassified optical images based on image collection protocol.}
    \label{fig:misclassified_optical}
\end{figure}

Table \ref{tab:mobilenet_results} suggests that thermography-based models are not sensitive to the image collection protocol, as the performance is stellar even with different lighting, camera distances, and patient poses. For the optical images, Figure \ref{fig:misclassified_optical} summarizes the errors associated with each of the 12 image collection protocols across the two cameras.
The plot suggests that having the knees stacked together and taking the camera from a further distance (50 cm) generally yielded better performance. The ring light setting also was less ideal for when the top knee was forward or backward as it may create more shadows.

\subsection*{Impact of Skin Tone for Erythema Detection}

To better understand which skin tone categories are most susceptible to misclassification for the 3 image types, we further investigated the performance for each of the four skin tone categories.
Table \ref{tab:erythema-categories-breakdown} summarizes the results of the test set for InterLow, Inter, InterMid, and InterHigh.
As can be seen from the Table, many of the images were from the InterMid skin tone category.
The results suggest that optical images perform slightly better for the Inter category (1 incorrect versus 2 for thermal imaging).
However, thermal imaging is better for the InterMid category as there are 4 fewer mistakes in thermal imaging suggesting that for darker skin tones it is harder to detect redness in the skin.

\begin{table}[t]
\floatconts
  {tab:erythema-categories-breakdown}%
  {\caption{Comparing classification performance of the 3 image types, optical (Opt.), B\&W thermal image, and color thermal image, on the erythema task by skin tone categories.}}%
{\vspace{-20pt} \footnotesize
  \begin{tabular}{l l l l l }
\toprule
\textbf{Skin tone}& \textbf{Category}& \textbf{Opt.} & \textbf{B\&W} & \textbf{Color} \\
\midrule
 \multirow{2}{*}{InterLow} & Correct &3  & 3& 3 \\
 & Incorrect & 1  & 1   &1 \\
 \midrule
\multirow{2}{*}{Inter} & Correct & 13 &12 & 12 \\
 &Incorrect & 1  & 2  &2 \\
 \midrule
  \multirow{2}{*}{InterMid} & Correct & 32 &36 & 36 \\
 &Incorrect & 8  & 4   & 4\\
 \midrule
 \multirow{2}{*}{InterHigh} & Correct& 5 &5 & 5 \\
 &Incorrect & 0  & 0  & 0\\
 
\bottomrule
\end{tabular}}
\end{table}

\subsection{Conclusion and Future work}
Our findings demonstrate the potential of thermography and CNNs to detect temperature changes in individuals with darker skin tones more reliably than optical images.
Although the MobileNetV2 performance is slightly lower for detecting erythema in slightly darker skin (InterMid skin tone category), the results are still better than their optical counterparts.
This suggests that thermography may be a viable solution for detecting PI even on dark skin tones -- a significant challenge under current clinical practices. 
Our preliminary results also suggest that thermography does not require adherence to a strict imaging protocol as there was little performance difference between the patient pose, lighting setting, and camera distance.
This may serve as a crucial finding as following a stringent imaging protocol may adversely impact nursing workflow and impede adoption. 

Our preliminary findings provide promising implications for the adoption of thermography for PI detection even for diverse skin tones. However, we do note that there are several limitations to our current work.
First, our patient sample is quite small (35 patients) with many from the InterMid skin tone category. 
As such, a larger sample size is necessary to draw more substantive conclusions.
Second, the images were collected in a controlled and simulated setting and did not include real PI.
Although there is some evidence that thermography is fairly robust to lighting, pose, and distance, this may change for real-time deployment as patients may not be capable of lying on their side or nurses may not provide high-quality images.
Third, we posit that our model can generalize across different body parts. However, to validate this hypothesis we need to expand our work to collect thermal imaging from more diverse locations.

\subparagraph{Acknowledgements}
This work was supported in part by the National Center for Advancing Translational Sciences of the National Institutes of Health under Award number UL1TR002378. The content is solely the responsibility of the authors and does not necessarily represent the official views of the National Institutes of Health.

\bibliography{jmlr-sample}

\appendix

\section{Dataset Details}\label{apd:data_collection}

Cupping was used to induce erythema. A total of 9 images, consisting of 1 control image and 8 other images taken from 0 to 7 minutes after cupping were captured per patient. At each timeframe, the erythema index was taken with a colorimeter. An erythema index of 6 color units above baseline is assigned a positive label, otherwise, negative. As a result, 57 (18\%) of the 314 images were labeled as positive.

Variations in lightning, distance, cameras, and posture was not conducted on the erythema dataset because the variation across individual responses to cupping is still not well understood. Moreover, a recent study \citep{bates2024thermal} found weak correlations between the erythema index and temperature changes when erythema was induced erythema on the forearm. Nonetheless, we plan to investigate the impact of different lighting conditions on erythema detection in darker skin tones in the future.

\begin{figure}[t]
    \centering
    \includegraphics[width=0.95\linewidth]{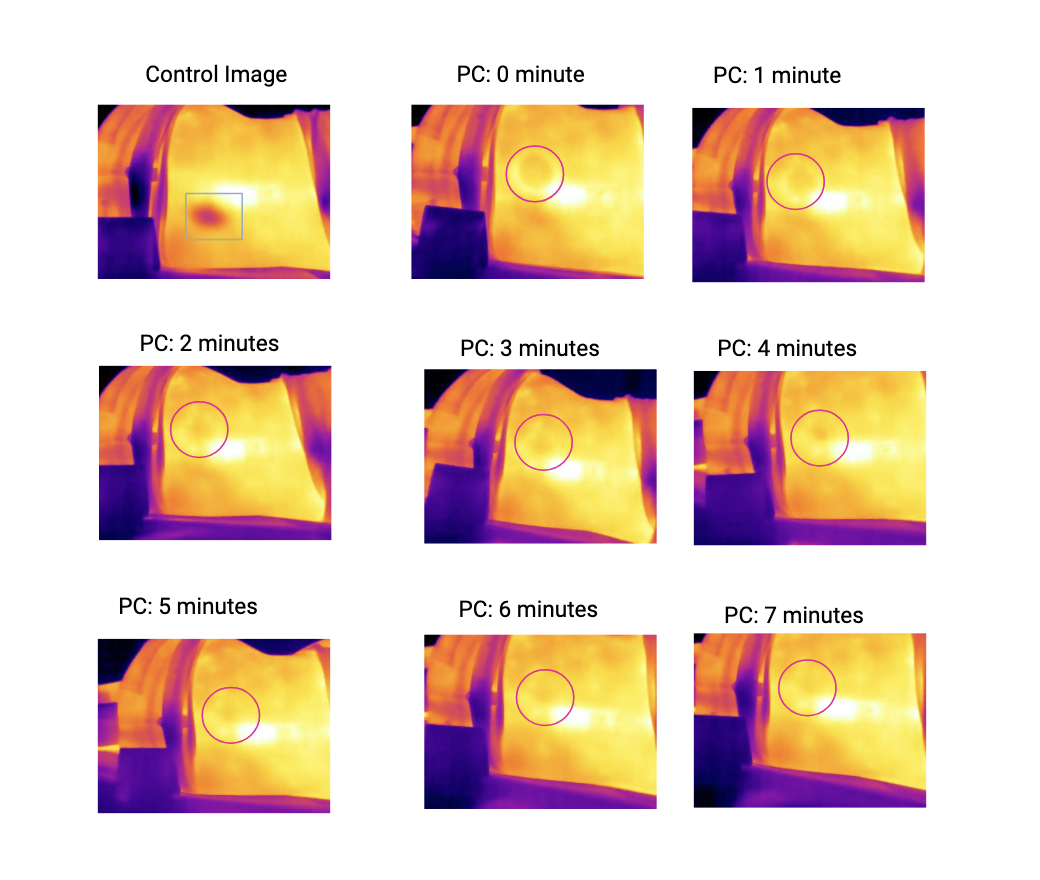}
    \vspace{-2ex}
    \caption{Progression of erythema in thermal image from $0-7$ minutes post-cupping. PC denotes Post-Cupping, square represents the cooling spot, circle represents the erythema spot.}
    \label{fig:erythema_progression}
\end{figure}

Figure \ref{fig:erythema_progression} shows the progression of erythema from 0 to 7 minutes. As is evident from the figure, as the time increases, the redness of the skin gradually fades away. While erythema is very clear post-cupping immediately (0 minutes), it is almost invisible at 7 minutes post-cupping.

Table \ref{tab:subject_distribution} shows the percentages and breakdowns by skin tone using the revised Eumelanin Human Skin Colour Scale categories \citep{sonenblum2023using} of the 35 subjects included in this study.

\begin{table}[t]
    \caption{Percentage distribution of subjects based on skin tone}
    \label{tab:subject_distribution}
    \centering
    \vspace{-0.5em}
    \footnotesize
    \begin{tabular}{l c}
     \toprule
      Modified Eumelanin Skin Tone Group & \% (n) \\
    \midrule
       Intermediate Low (InterLow)  & 11.4 (4) \\
       Intermedidate (Inter)  &17.1 (6) \\
       Intermediate Mid (InterMid) & 62.9 (22)  \\
        Intermediate High (InterHigh) & 8.6 (3)\\
        \bottomrule
    \end{tabular}
\end{table}

\section{Additional Methods}\label{apd:methods}

\begin{figure}[t]
    \centering
\includegraphics[width=1\linewidth]{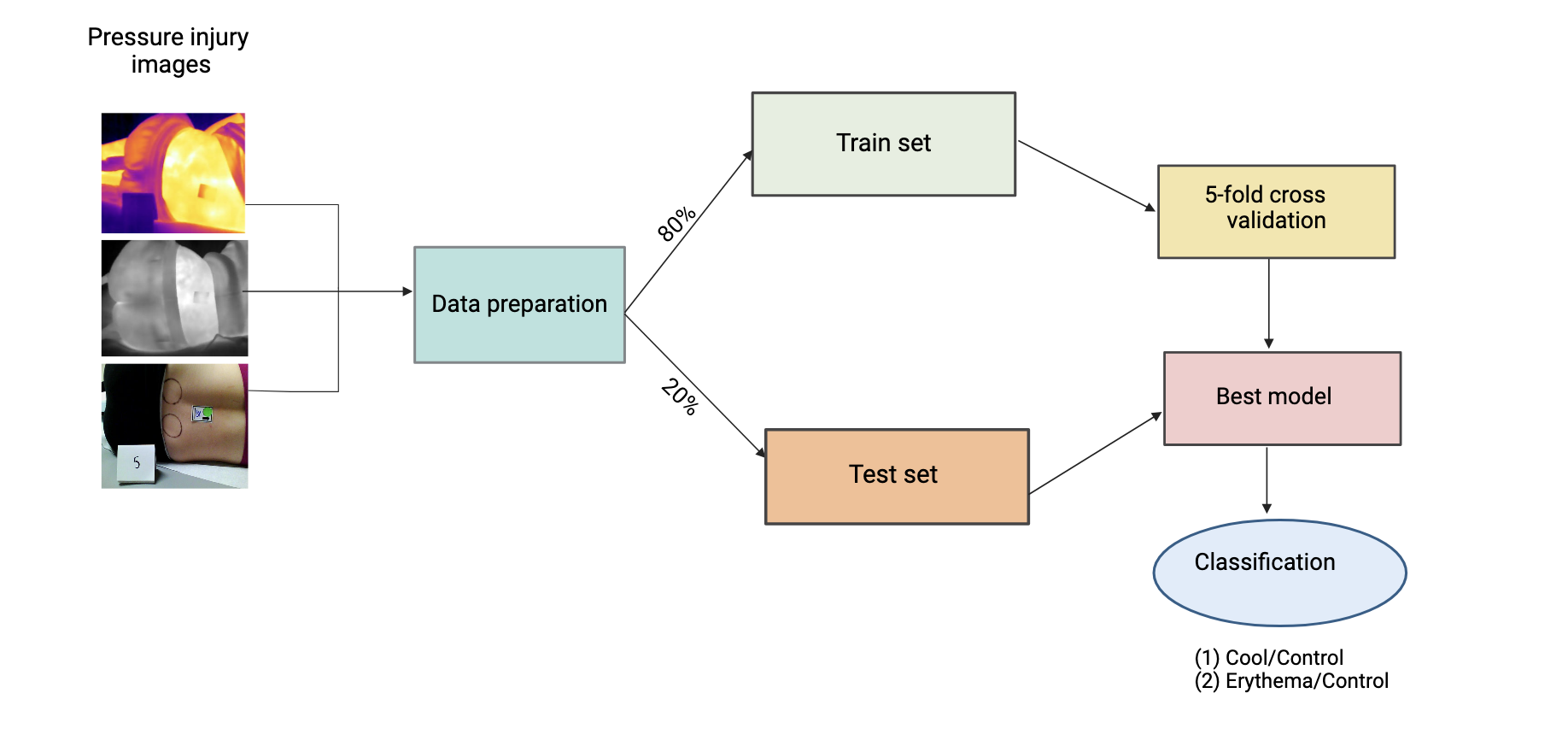}
\vspace{-2em}
    \caption{Workflow for classifying cooling or erythema images.}
    \label{fig:model}
\end{figure}

Data is split randomly into training and test sets with 80\% for training and 20\% for testing as shown in Table \ref{tab:sample}. For fair comparison, we ensure the same images appear in the train and test splits across all image modalities. For instance, if image 1 is in the training data of the optical dataset, the corresponding image 1 from both the thermal colored dataset and the thermal B\&W dataset will be included in their respective training data sets. After data splitting, we perform 5-fold cross-validation using the train data, with 4 folds for training and 1 fold for validation as shown in Figure \ref{fig:model}. 
Both train and test images were normalized with mean of [0.485, 0.456, 0.406] and standard deviation of [0.229, 0.224, 0.225]. The best model based on the highest validation accuracy over the 5-fold cross-validation was selected and finally used to classify the images in the test set.

\begin{table}[t]
\floatconts
  {tab:sample} 
  {\caption{Number of train and test samples for Erythema and Cool/Base images}} 
  {\centering 
  \vspace{-1.5em}
  \footnotesize
  \begin{tabular}{l p{3.4cm}cc}
    \toprule
    & \textbf{\# Samples} & \textbf{Train} & \textbf{Test} \\
    \midrule
    Cooling & Optical, B\&W thermal, Color thermal & 1337 & 343 \\
    \addlinespace
    Erythema & Optical, B\&W thermal, Color thermal & 251 & 63 \\
    \bottomrule
  \end{tabular}
  }
\end{table}

\subsection*{Model Hyperparameters }
We used the MobileNetV2,which is pre-trained on ImageNet \citep{russakovsky2015imagenet} for fine-tuning to speed up the model convergence and improve the overall classification accuracy.
Hyperparameters used for training are (a) batch size of 32 (b) 50 epochs (c) learning rate of \(1\times e^{-3}\) and (d) early stopping (patience of 10). Adam optimizer \citep{kingma2020method} and binary-cross entropy loss function were used. The Adam optimizer is a stochastic gradient descent approach that uses adaptive estimates of first-order and second-order moments. 
The best model based on the highest validation accuracy was saved and used for classification on the test set.

Our system was implemented using Python, leveraging the PyTorch framework \citep{paszke2019pytorch} for deep learning operations. Additional libraries used are  Numpy \citep{harris2020array} for numerical computations, Matplotlib \citep{hunter2007matplotlib} for data visualization, and OpenCV \citep{bradski2000opencv} for image processing tasks. The development was carried out using Jupyter notebook within Visual Studio Code.

\section{Additional Results}\label{apd:results}

\subsection*{Classification Analysis}
\begin{figure*}[ht]
    \centering
    \includegraphics[width= 0.8\textwidth,height=13em]{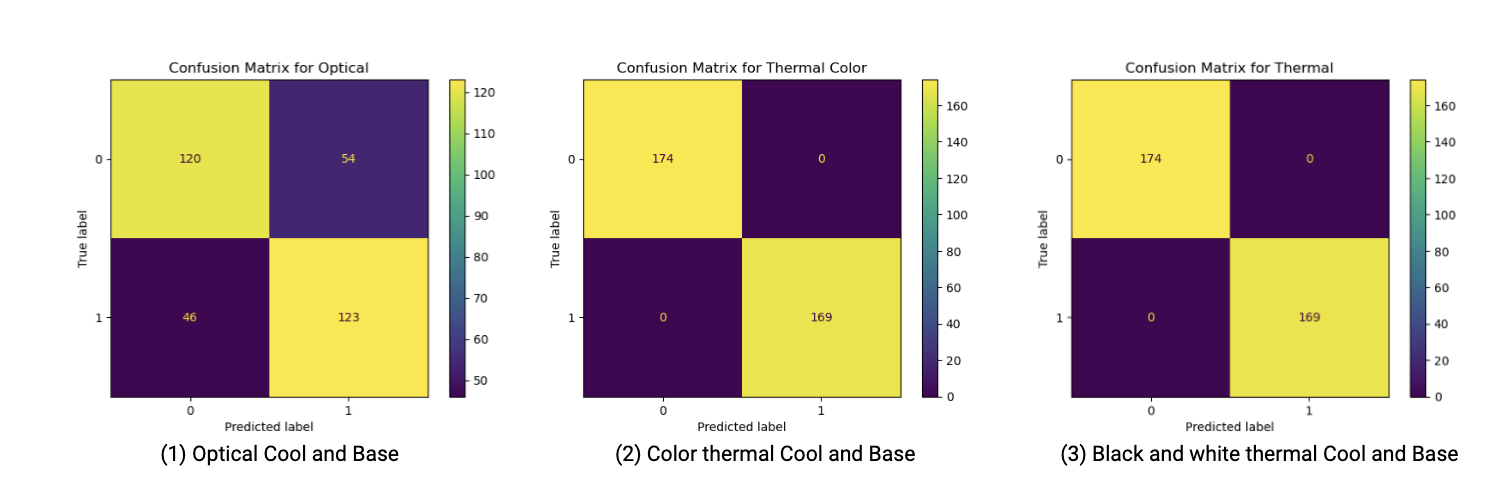}
    \vspace{-12pt}
    \caption{Confusion matrix of the three image modalities in the cooling classification task.}
    \label{fig:confusion_cool_base}
\end{figure*}

\begin{figure*}[ht]
    \centering    \includegraphics[width=0.8\textwidth,height=13em]{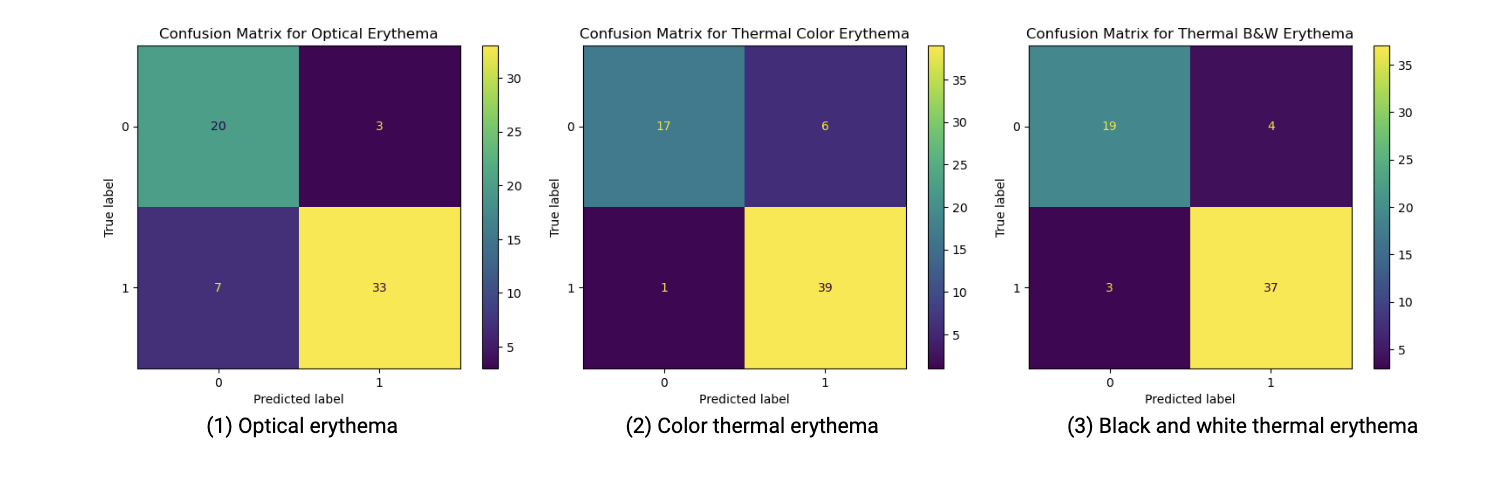}
    \vspace{-12pt}
    \caption{Confusion matrix of the three imaging modalities used in the erythema classification task.}
    \label{fig:confusion_erythema}
\end{figure*}

Figures \ref{fig:confusion_cool_base} and \ref{fig:confusion_erythema} show the confusion matrix for the cooling dataset and the erythema dataset respectively, which offers insights into the classification performance of our fine-tuned MobileNetV2 model across the different imaging modalities. These visual representations allow us to assess the accuracy and potential misclassifications.
With the cooling dataset (Figure \ref{fig:confusion_cool_base}), both thermal color and thermal black and white imaging demonstrate perfect classification (174 true negatives and 169 true positives).

The erythema dataset is also examined, comparing the model's predictions against true labels for the presence or absence of skin redness. Among these, the thermal color imaging shows the highest overall accuracy, with 56 correct predictions (17 true negatives and 39 true positives) as shown in Figure \ref{fig:confusion_erythema}. 

\begin{figure}[ht]
    \centering
\includegraphics[width=0.75\linewidth]{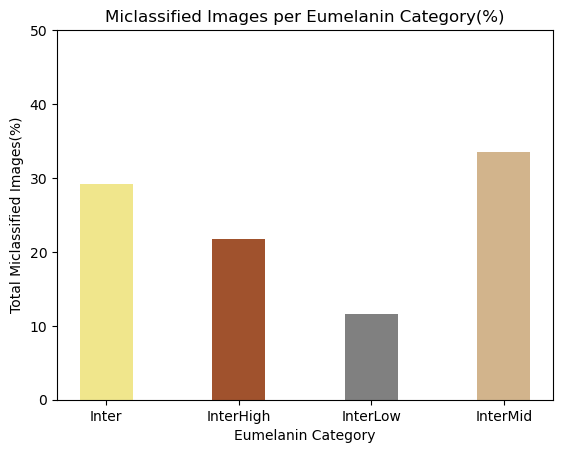}
\vspace{-1em}
    \caption{Eumelanin categories of misclassified optical cooling images.}
    \label{fig:eumelanin_category}
\end{figure}

\subsection{Additional Image Protocol Analysis}
We also assessed the protocols used in the data collection process to identify which will potentially be useful for detecting PI.
We analyzed the misclassified optical images in the cooling dataset, considering the number of misclassified images per protocol used. In Figure \ref{fig:misclassified_optical}, each protocol is represented as Combo. A higher number of images is an indication that a given protocol will perform poorly in detecting pressure injury, while a lower number of images indicates a potentially high performance of this protocol in detecting pressure injury.
These 12 protocols represent distinct combinations, each varying by knee position (Forward, Stacked, or Back), lighting condition (Room Light or Ring Light), and distance (35 cm or 50 cm). Analysis reveals that Combo 3 (Knee Forward, Ring Light, 35 cm) resulted in the highest number of misclassifications (12-13), while Combo 8 (Knees Stacked, Ring Light, 50 cm) had the lowest (about 5). As shown in Figure \ref{fig:misclassified_optical}, misclassification rates varied considerably, ranging from 5 to 13 across different combinations. Generally, the 35 cm distance led to more misclassifications compared to 50 cm under similar conditions. The effects of lighting and knee position on misclassification rates were mixed, showing no consistent pattern. These findings suggest that imaging conditions significantly impact classification accuracy, with certain setups posing greater challenges for the algorithm. It also shows that images taken with knees stacked under ring lighting conditions at a distance of 50cm have a high potential of being correctly classified by deep learning algorithms.




\end{document}